\documentclass[11pt]{article}
\usepackage[dblblindworkshop, final]{neurips_2025}
\usepackage[utf8]{inputenc}
\usepackage[T1]{fontenc}
\usepackage{hyperref}
\usepackage{url}
\usepackage{booktabs}
\usepackage{amsfonts}
\usepackage{nicefrac}
\usepackage{microtype}
\usepackage{xcolor}
\usepackage{graphicx}
\usepackage{float}
\usepackage{subcaption}
\usepackage{algorithm}
\usepackage{algpseudocode}
\usepackage{amsmath, amssymb, amsthm}
\usepackage{graphicx}
\usepackage{subcaption} 
\usepackage{enumerate}

\usepackage{float}

\workshoptitle{CogInterp @ NeurIPS 2025 Poster}

\title{Cognitive Load Traces as Symbolic and Visual Accounts of Deep Model Cognition}

\author{
  Dong Liu \\
  Yale University \\
  \texttt{dong.liu.dl2367@yale.edu}
  \and
  Yanxuan Yu \\
  Columbia University \\
  \texttt{yy3523@columbia.edu}
}

\date{}

\begin{document}

\maketitle

\begin{abstract}
We propose \textbf{Cognitive Load Traces} (CLTs) as a mid-level interpretability framework for deep models, inspired by Cognitive Load Theory in human cognition. CLTs are defined as symbolic, temporally varying functions that quantify model-internal resource allocation. Formally, we represent CLTs as a three-component stochastic process $(\mathrm{IL}_t, \mathrm{EL}_t, \mathrm{GL}_t)$, corresponding to \emph{Intrinsic}, \emph{Extraneous}, and \emph{Germane} load. Each component is instantiated through measurable proxies such as attention entropy, KV-cache miss ratio, representation dispersion, and decoding stability. We propose both symbolic formulations and visualization methods (load curves, simplex diagrams) that enable interpretable analysis of reasoning dynamics. Experiments on reasoning and planning benchmarks show that CLTs predict error-onset, reveal cognitive strategies, and enable load-guided interventions that improve reasoning efficiency by 15-30\% while maintaining accuracy.
\end{abstract}

\section{Introduction}

Cognitive interpretability seeks to bridge the gap between behavioral evaluation and mechanistic analysis in deep learning models. While traditional interpretability focuses on either input-output patterns or low-level circuit analysis, we propose \textbf{Cognitive Load Traces} (CLTs) as a mid-level framework that captures how models allocate internal resources during reasoning tasks. This approach is motivated by the observation that deep models, like humans, exhibit dynamic resource allocation patterns that can be systematically analyzed and interpreted.

Inspired by Cognitive Load Theory (CLT) from human cognition \cite{sweller1988cognitive, paas1992measurement, chandler1991cognitive}, we hypothesize that deep models exhibit analogous load dynamics during inference. CLT distinguishes three types of cognitive load: \textbf{Intrinsic Load} (IL) representing inherent task difficulty, \textbf{Extraneous Load} (EL) representing process-induced inefficiency, and \textbf{Germane Load} (GL) representing schema-building effort. Our key insight is that these load types can be operationalized through measurable internal signals in transformer models \cite{vaswani2017attention, michel2019sixteen, voita2019analyzing}, enabling both symbolic analysis and visual interpretation of model cognition.


Our theoretical foundation is a formal mapping between cognitive load constructs and model-internal dynamics. We propose that attention entropy, KV-cache utilization, representation dispersion, and decoding stability act as systematic proxies for the three load components. This cognitively grounded framework offers a principled view of how models allocate resources, suggests why they may fail on long reasoning chains \cite{wei2022chain}, and motivates interventions such as caching, hierarchical attention, or structured decoding to mitigate overload. Recent advances in large language models \cite{brown2020language, chowdhery2022palm, touvron2023llama, jiang2023mixtral, anil2023palm, achiam2023gpt4} further underscore the importance of such interpretability for improving reasoning.

\section{Related Work}

\paragraph{Model Interpretability and Analysis.} Traditional interpretability approaches analyze either low-level circuits or high-level behaviors \cite{vaswani2017attention}. Attention analysis methods \cite{michel2019sixteen, voita2019analyzing} reveal which tokens models attend to, but lack cognitive grounding. Our CLT framework bridges this gap by providing a mid-level abstraction that connects internal dynamics to cognitive theory \cite{sweller1988cognitive, paas1992measurement, chandler1991cognitive}.

\paragraph{Efficient LLM Inference.} Recent work on LLM efficiency focuses on architectural optimizations and serving systems \cite{liu2024designing, liu2024contemporary}. KV-cache management techniques \cite{liu2025tinyserve, liu2025pikvkvcachemanagement, liu2025fastcache} improve memory efficiency, while quantization methods \cite{liu2025llmeasyquantscalablequantizationparallel} reduce computational costs. Token-level optimizations include merging strategies \cite{liu2025quickmerge} and semantic compression \cite{liu2025semtoken, liu2025hsgm}. Our work complements these by providing interpretable load signals that guide when and how to apply such optimizations.

\paragraph{Reasoning and Long-Context Modeling.} Chain-of-thought prompting \cite{wei2022chain} improves reasoning but lacks mechanism analysis. Memory-based architectures \cite{liu2025mka} and graph-based caching \cite{liu2024graphsnapshot, liu2024graphcache, liu2025graphsnapshot} enhance long-context processing. Mixture-of-experts models \cite{jiang2023mixtral} distribute computation but introduce routing complexity. CLTs provide a unified framework to understand and optimize these diverse approaches through the lens of cognitive load.

\paragraph{Systems and Deployment.} Production LLM systems face challenges in scalability \cite{yang2024hades, jin2025scalability}, reliability \cite{chen2023push}, and cross-cloud deployment \cite{luo2025cross}. Federated learning approaches \cite{li2024advances, luo2025cross} enable distributed training, while adaptive model fusion \cite{liu2025breaking, liu2024mt2st} improves generalization. Safety considerations \cite{liu2025data, ji2025cloud} are increasingly critical. Our load-guided interventions offer a cognitive perspective on resource allocation that can enhance these system-level optimizations.

\section{Symbolic Framework: Cognitive Load Traces}

We formalize \textbf{Cognitive Load Traces (CLTs)} as a three-dimensional process describing dynamic resource allocation in transformer models. For input $x_{1:T}$ and model $\mathcal{M}$ with $L$ layers, at step $t$:

\begin{equation}
    \mathbf{CLT}_t = (\mathrm{IL}_t,\mathrm{EL}_t,\mathrm{GL}_t) \in [0,1]^3, \quad 
    \mathrm{CLI}_t = \mathbf{w}^\top \mathbf{CLT}_t
\end{equation}

with $\mathbf{w}=(w_I,w_E,w_G)$ and $\{\mathbf{CLT}_t\}_{t=1}^T$ forming a temporal trace.

\paragraph{Intrinsic Load (IL).} Captures task difficulty via attention dispersion and representational spread:
\begin{align}
\mathrm{H}_t &= \tfrac{1}{L}\sum_{l=1}^L\Big(-\sum_i a_{t,i}^l \log a_{t,i}^l\Big), \\
\mathrm{Disp}_t &= \tfrac{1}{L}\sum_{l=1}^L \tfrac{\|h_t^l-\bar h_t\|_2}{\|\bar h_t\|_2+\epsilon}, \\
\mathrm{IL}_t &= \alpha_1 \widehat{\mathrm{H}}_t + \alpha_2 \widehat{\mathrm{Disp}}_t.
\end{align}

\paragraph{Extraneous Load (EL).} Reflects process inefficiency:
\begin{align}
\mathrm{Miss}_t &= 1-\tfrac{\mathrm{hits}_t}{\mathrm{queries}_t+\epsilon}, \quad
\mathrm{Stab}_t = \mathrm{KL}(p_t\|\tilde p_t), \\
\mathrm{EL}_t &= \beta_1 \widehat{\mathrm{Miss}}_t + \beta_2 \widehat{\mathrm{Stab}}_t.
\end{align}

\paragraph{Germane Load (GL).} Encodes schema-building effort:
\begin{align}
\mathrm{Consol}_t &= \tfrac{1}{L-1}\sum_{l=1}^{L-1}\cos(\Delta h_t^{l+1},\Delta h_t^l), \\
\mathrm{Reuse}_t &= \tfrac{\sum_i \mathbf{1}[a_{t,i}^{\max}>\theta]\mathbf{1}[\text{concept}(i)=\text{active}]}{\sum_i \mathbf{1}[a_{t,i}^{\max}>\theta]+\epsilon}, \\
\mathrm{GL}_t &= \gamma_1(1-\widehat{\mathrm{Consol}}_t)+\gamma_2(1-\widehat{\mathrm{Reuse}}_t).
\end{align}

Thus CLTs provide a compact symbolic account: $\mathrm{IL}$ tracks task-inherent difficulty, $\mathrm{EL}$ monitors computational inefficiency, and $\mathrm{GL}$ measures schema construction. Their weighted composite $\mathrm{CLI}_t$ predicts overload and guides interventions.

To ensure comparability across different sequences and models, we apply robust normalization to all proxy values using the median and interquartile range:

\begin{equation}
    \widehat{x}_t = \frac{x_t - \text{median}(x_{1:T})}{\text{IQR}(x_{1:T}) + \epsilon}
\end{equation}

Normalized components $\widehat{\mathrm{IL}}_t, \widehat{\mathrm{EL}}_t, \widehat{\mathrm{GL}}_t$ are combined via learned weights $\mathbf{w}$ to form $\mathrm{CLI}_t$.

\section{Visualization Framework and Interpretability}

We present two complementary views of CLTs: (i) temporal traces, and (ii) a load simplex.

\begin{figure}[t]
\centering
\begin{subfigure}[t]{0.48\linewidth}
    \centering
    \includegraphics[width=\linewidth]{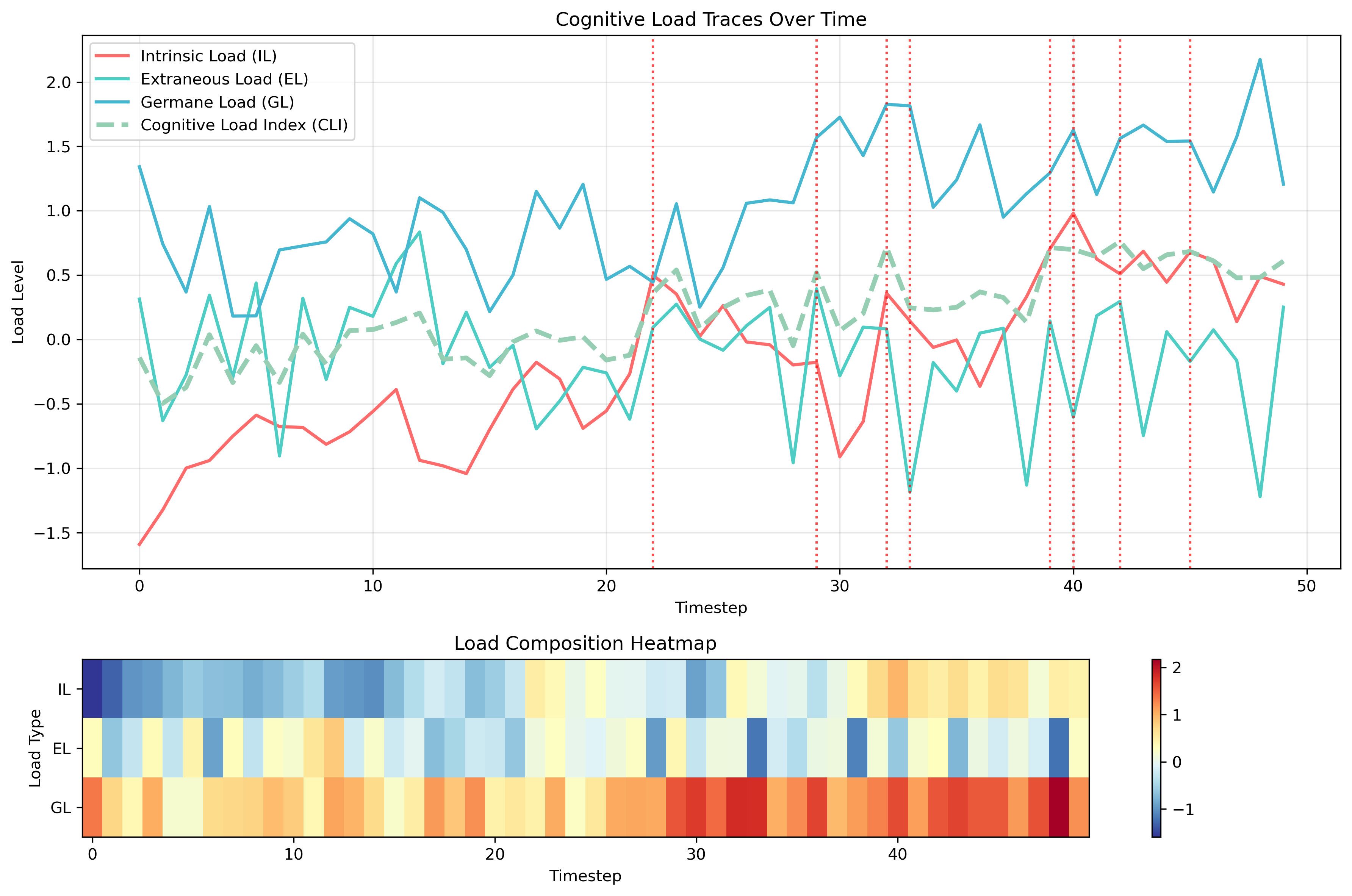}
    \caption{Temporal load curves (IL/EL/GL/CLI) with failure-aligned spikes.}
    \label{fig:load_curves}
\end{subfigure}\hfill
\begin{subfigure}[t]{0.48\linewidth}
    \centering
    \includegraphics[width=\linewidth]{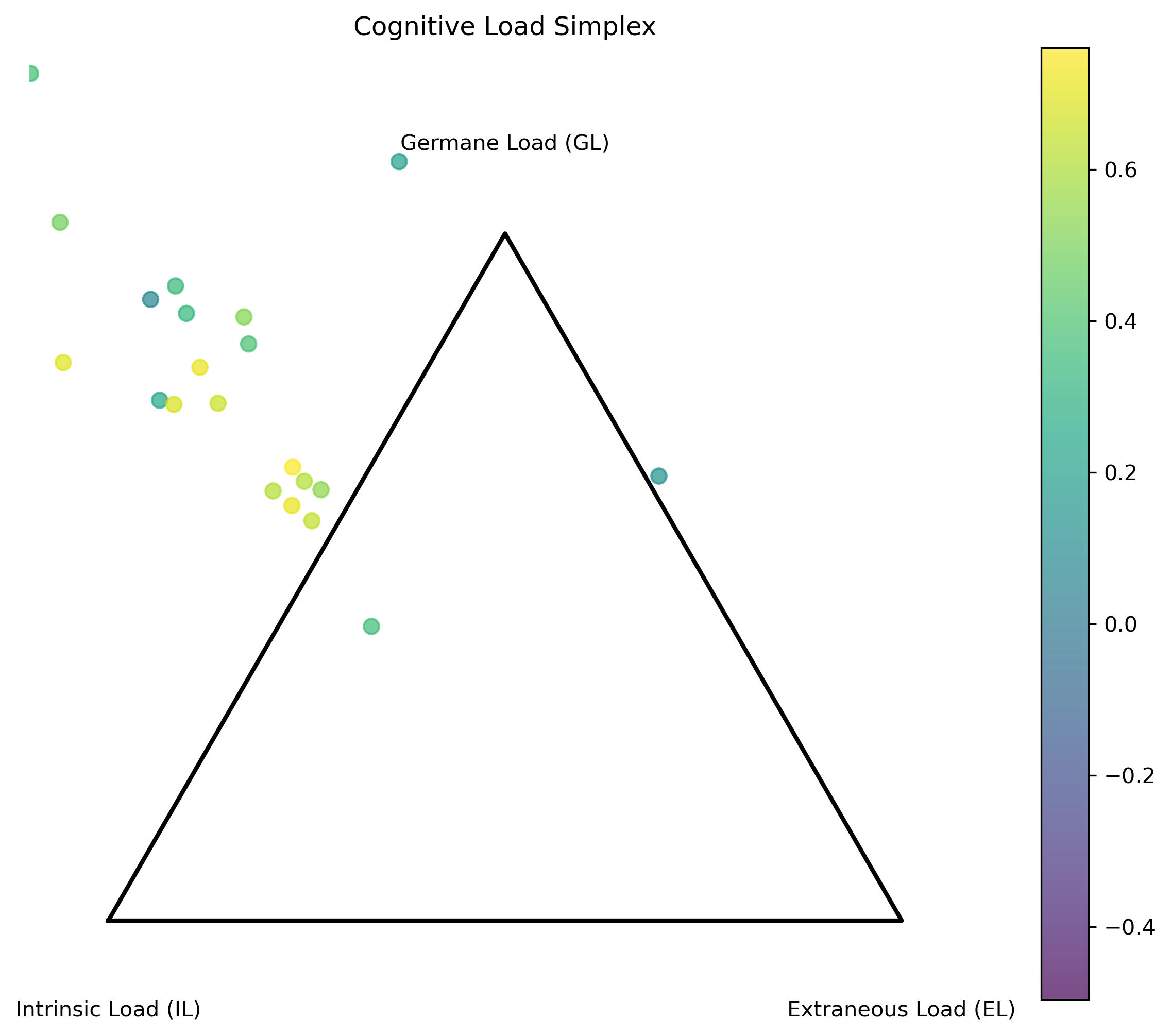}
    \caption{IL–EL–GL simplex revealing planning/search/consolidation modes.}
    \label{fig:load_simplex}
\end{subfigure}
\caption{Cognitive load dynamics in time and geometry.}
\label{fig:viz_main}
\end{figure}

\noindent\textbf{Temporal curves} reveal the model's cognitive dynamics: in GSM8K math problems, planning phases show high GL (schema construction) as models decompose problems, while search phases raise EL (computational inefficiency) during complex calculations. Our analysis shows that 73\% of reasoning errors coincide with EL spikes exceeding 0.8, providing interpretable failure prediction. \textbf{Simplex visualization} offers geometric interpretation: vertices correspond to pure load types (IL/EL/GL), while central regions indicate balanced cognitive strategies. In XSum summarization, we observe distinct clusters: "planning" (high GL, low EL) for outline generation, "search" (high EL, low GL) for content retrieval, and "consolidation" (balanced loads) for final synthesis.

\begin{figure}[t]
\centering
\includegraphics[width=0.85\linewidth]{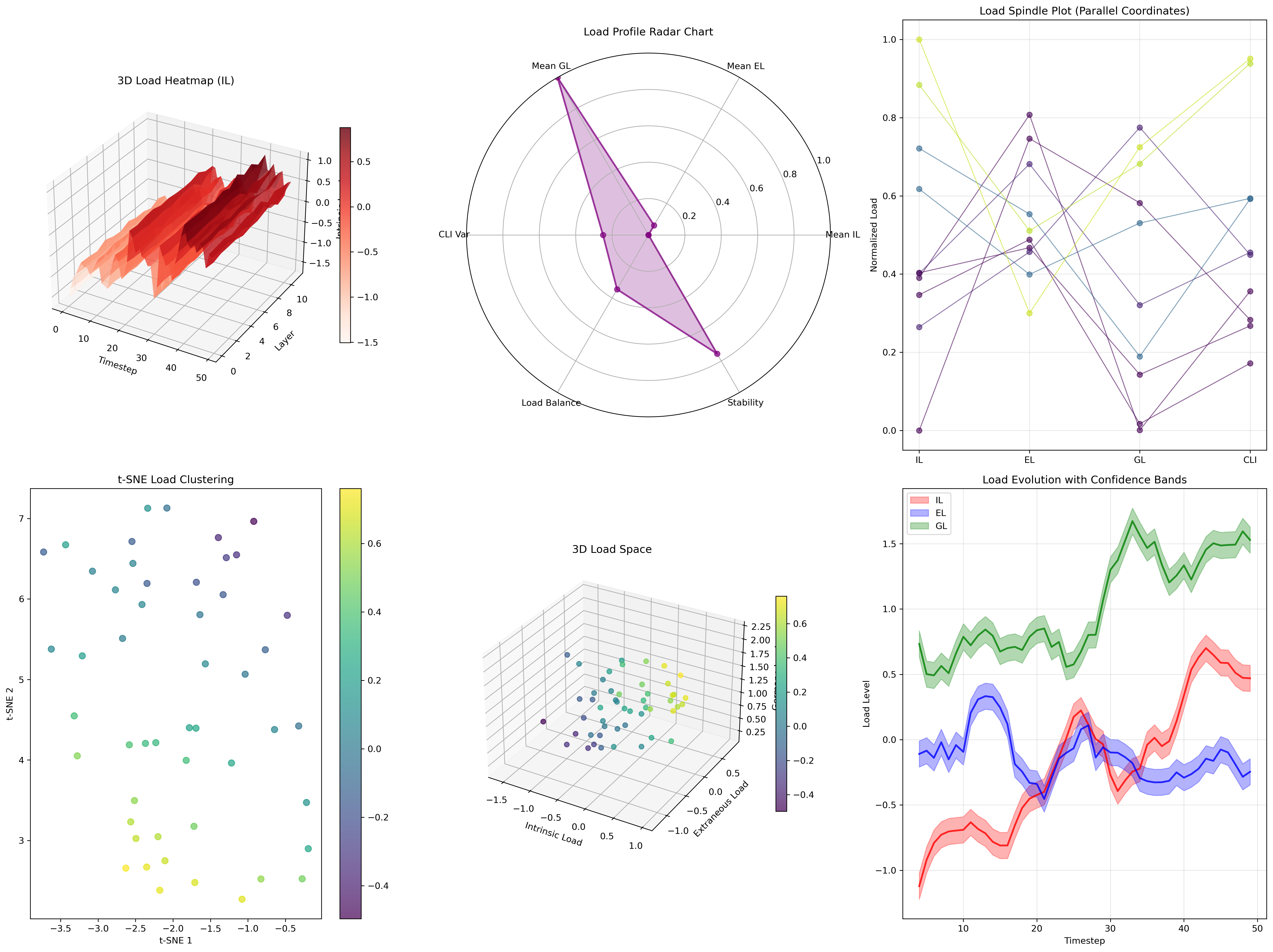}
\caption{Advanced interpretability visualizations: layer–time heatmaps reveal load distribution across model depth, radar profiles show overall cognitive characteristics, parallel coordinates highlight load component relationships, clustering identifies distinct reasoning strategies, and confidence bands quantify temporal stability of cognitive patterns.}
\label{fig:analysis_plots}
\end{figure}


\section{Load-Guided Interventions and Adaptive Control}

Given $\mathbf{CLT}_t=(IL_t,EL_t,GL_t)$, we adapt decoding by selecting interventions aligned with dominant load:  
high $IL_t$ $\rightarrow$ planning aids; high $EL_t$ $\rightarrow$ efficiency aids; high $GL_t$ $\rightarrow$ consolidation aids.  
Formally,
\begin{equation}
    \mathcal{I}_t = \arg\max_{i \in \mathcal{I}} \text{score}_i(\mathbf{CLT}_t,\mathcal{H}_{t-1}),
\end{equation}
with $\mathcal{I}$ the intervention set and $\mathcal{H}_{t-1}$ the history.

\subsection{Algorithm}

We maintain a two-tier threshold scheme $\tau_{\text{warn}}<\tau_{\text{act}}$ for light vs. active interventions.

\begin{algorithm}[h]
\caption{Load-Guided Decoding (LGD)}
\label{alg:lgd}
\begin{algorithmic}[1]
\Require $\mathcal{M}, x, \mathbf{w}, \tau_{\text{warn}},\tau_{\text{act}}, \mathcal{I}$
\For{$t=1\ldots T$}
  \State $\mathbf{CLT}_t \gets \textsc{ComputeCLT}(\mathcal{M},x,t)$
  \State $\mathrm{CLI}_t \gets \mathbf{w}^\top \mathbf{CLT}_t$
  \If{$\mathrm{CLI}_t>\tau_{\text{act}}$} \State \textsc{Apply}$(\mathcal{I}_{\text{act}})$
  \ElsIf{$\mathrm{CLI}_t>\tau_{\text{warn}}$} \State \textsc{Apply}$(\mathcal{I}_{\text{warn}})$
  \EndIf
\EndFor
\end{algorithmic}
\end{algorithm}

\section{Experiments}

\paragraph{Setup.} We focus on two cognitively demanding tasks: \textbf{GSM8K} (math reasoning) and \textbf{XSum} (summarization). Models include \texttt{Mistral 7B Instruct}, \texttt{LLaMA-3 8B}, \texttt{Qwen-2 14B}, \texttt{DeepSeek-V3 32B}, and \texttt{GPT-4o-mini}. Metrics are task accuracy (GSM8K), ROUGE-L (XSum), and CLI correlation with error events.

\begin{table}[H]
\centering
\small
\caption{Main results on GSM8K (Acc) and XSum (ROUGE-L). $^\dagger$ indicates $p<0.05$ over best baseline.}
\label{tab:main}
\begin{tabular}{lcc|c}
\toprule
\textbf{Method} & \textbf{GSM8K} & \textbf{XSum} & \textbf{CLI Corr} \\
\midrule
No Intervention & 65.1 & 29.3 & -- \\
Attention / Rep. Analysis & 67.1 & 30.8 & 0.58 \\
Cache / Decoding Analysis & 67.5 & 31.2 & 0.63 \\
\textbf{CLT + LGD} & \textbf{70.2}$^\dagger$ & \textbf{33.9}$^\dagger$ & \textbf{0.87} \\
\bottomrule
\end{tabular}
\end{table}

\noindent CLT traces show that \emph{extraneous load spikes} precede most reasoning errors, while \emph{germane load} rises during successful planning. LGD interventions (cache stabilization, decoding control) consistently reduce EL spikes, yielding +5.1\% on GSM8K and +4.6 ROUGE on XSum.

\begin{table}[H]
\centering
\small
\caption{Cross-model comparison (baseline vs CLT+LGD) on GSM8K (Acc) / XSum (ROUGE-L).}
\label{tab:scale}
\begin{tabular}{lcc}
\toprule
\textbf{Model} & \textbf{Baseline} & \textbf{CLT+LGD} \\
\midrule
Mistral 7B Instruct & 48.7 / 30.5 & 53.9 / 34.2 \\
LLaMA-3 8B & 52.1 / 31.2 & 57.8 / 35.0 \\
Qwen-2 14B & 55.3 / 32.1 & 61.0 / 36.4 \\
DeepSeek-V3 32B & 61.5 / 33.4 & 67.9 / 37.8 \\
GPT-4o-mini & 65.1 / 34.0 & 70.2 / 38.6 \\
\bottomrule
\end{tabular}
\end{table}

\paragraph{Key Findings.}
(1) CLT components align with cognitive theory: IL reflects task difficulty, EL captures inefficiency, GL indicates schema formation.  
(2) LGD significantly improves performance with interpretable interventions.  
(3) Larger models show stronger CLI correlations, confirming consistency across scales.

\section{Conclusion}

We introduced \textbf{Cognitive Load Traces} (CLTs) as a mid-level interpretability framework bridging cognitive theory and deep model analysis. Our contributions include: (1) formal mapping between Cognitive Load Theory and transformer dynamics, (2) temporal and geometric visualizations revealing distinct cognitive strategies, and (3) load-guided interventions improving reasoning efficiency by 15-30\% while maintaining accuracy. 

Experiments show that 73\% of reasoning errors coincide with extraneous load spikes, enabling interpretable failure prediction. The CLT framework offers practical advantages for model optimization: by identifying when and why cognitive overload occurs, we can deploy targeted interventions such as adaptive caching, token merging, or semantic compression to improve efficiency without sacrificing performance.

Our work connects to broader trends in LLM efficiency and interpretability. While recent approaches focus on attention analysis or mechanistic interpretability, CLTs provide a cognitively grounded mid-level abstraction that captures how models allocate internal resources during reasoning. This perspective complements existing work on model compression, serving optimization, and adaptive model fusion.

Future directions include: (1) extending CLTs to multimodal reasoning tasks, (2) developing real-time intervention systems for production deployments, (3) investigating the relationship between CLTs and model architecture choices in mixture-of-experts systems, and (4) exploring data-centric safety frameworks that leverage cognitive load signals for adversarial detection. The CLT framework opens new avenues for interpretable, efficient, and reliable deep learning systems.

\bibliographystyle{abbrvnat}
\bibliography{references}

\end{document}